# Biomolecular Analysis of Soil Samples and Rock Imagery for Tracing Evidence of Life Using a Mobile Robot

1st Shah Md Ahasan Siddique
*Department of Mechanical Engineering*
*Military Institute of Science and Technology*
Dhaka, Bangladesh
ahasan.sid42@gmail.com

2nd Ragib Tahshin Rinath
*Department of Mechanical Engineering*
*Military Institute of Science and Technology*
Dhaka, Bangladesh
ragibtahshin@gmail.com

3rd Shakil Mosharrof
*Department of Computer Science & Engineering*
*Military Institute of Science and Technology*
Dhaka, Bangladesh
shakilmrf8@gmail.com

4th Syed Tanjib Mahmud
*Department of Mechanical Engineering*
*Military Institute of Science and Technology*
Dhaka, Bangladesh
ucchash123@gmail.com

5th Sakib Ahmed
*Department of Mechanical Engineering*
*Military Institute of Science and Technology*
Dhaka, Bangladesh
ahmedsakib18030@gmail.com

*Abstract*—The search for evidence of past life on Mars presents a tremendous challenge that requires the usage of very advanced robotic technologies to overcome it. Current digital microscopic imagers and spectrometers used for astrobiological examination suffer from limitations such as insufficient resolution, narrow detection range, and lack of portability. To overcome these challenges, this research study presents modifications to the Phoenix rover to expand its capability for detecting biosignatures on Mars. This paper examines the modifications implemented on the Phoenix rover to enhance its capability to detect a broader spectrum of biosignatures. One of the notable improvements comprises the integration of advanced digital microscopic imagers and spectrometers, enabling high-resolution examination of soil samples. Additionally, the mechanical components of the device have been reinforced to enhance maneuverability and optimize subsurface sampling capabilities. Empirical investigations have demonstrated that Phoenix has the capability to navigate diverse geological environments and procure samples for the purpose of biomolecular analysis. The biomolecular instrumentation and hybrid analytical methods showcased in this study demonstrate considerable potential for future astrobiology missions on Mars. The potential for enhancing the system lies in the possibility of broadening the range of detectable biomarkers and biosignatures.

## I. INTRODUCTION

In the course of human history, scientists have persistently attempted to push the limits of technological progress in their quest to decipher the mysterious complexities of our planet's environment and investigate the potential of life elsewhere in the universe. Within the context of this quest, the employment of mobile robotic platforms has emerged as a promising strategy for the evaluation of rock images and the performance of biomolecular investigations on soil samples. This study paper is a key milestone in the ongoing scientific investigation since it gives an in-depth analysis of the changes and additions that were implemented on our rover, Phoenix, with the intention of increasing the capabilities it possesses. Consequently, this report is a noteworthy achievement.

It is impossible to exaggerate the significance of improving our robotic systems since these vehicles act as our representatives in far-flung and frequently dangerous regions. An continual dedication to advance the present state of the art has been the driving force behind the significant amount of effort that has been spent to the process of reevaluating and improving Phoenix's functions. This lengthy endeavour covers a myriad of diverse stages, some of which include the incorporation of cutting-edge digital sensors as well as advancements to mechanical components. Intricate biomolecular research will be easier for the rover to carry out as a result of this.

In the forthcoming sections of this publication, we will delve into the intricacies of these momentous advancements. This will encompass the technological complexities as well as the methodologies that were employed to elevate the scientific scrutiny of Phoenix to an unprecedented echelon. Our research possesses the potential to not only deepen our comprehension of terrestrial ecosystems but also shed illumination on the quandary of whether or not life thrives elsewhere in the universe. This study aims to yield a substantial contribution to the rapidly evolving realm of astrobiology and establish the foundation for future endeavors that may one day facilitate the determination of whether or not other worlds are amenable to human habitation.

So,the novelty of this paper is as follows :

1) The incorporation of a biomolecular analytics subsystem onto a functional and cost-efficient rover platform.
2) Mechanical improvements for accessing a broader variety of samples, such as new actuators and suction pumps, permit access to a wider variety of biomarkers.
3) The hybrid approach for analysis of biomolecules that combines multiple techniques.

## II. RELATED WORKS

Extensive research has been conducted to explore various ways for identifying potential biosignatures and biomarkers in the pursuit of finding evidence of past or current life on Mars. The research performed by Smith et al. [1] primarily centred around the analysis of ground ice samples from Mars. The objective was to identify certain chemical compounds that might potentially serve as indicators of biological activity. Additionally, the researchers investigated several climatic parameters that are relevant to the potential habitability of Mars. The expedition that spanned 90 sols yielded valuable insights about biomarkers, although with certain limitations in its reach. In a similar vein of inquiry, Jones et al. [2] conducted a survey of Mars analogue settings on Earth in order to investigate biosignatures. Through the examination of organic molecules, liquid water activity, and energy sources in various harsh environments, the observational research has established rigorous criteria for the detection of indications of life under demanding circumstances. Nevertheless, terrestrial places just offer a limited representation of the Mars environment. In addition to conducting observational research, Zaman et al. [3] developed a quick life detection methodology that enables autonomous analysis of samples through the utilisation of several biomolecules. Although the experimental prototype has demonstrated promising results in predicting sample categories, it would need more optimisation to be suitable for real-world applications within the limits of a rover. Similarly, Pacelli et al. [4] conducted a series of verification experiments in which extremophile organisms were subjected to simulated Mars conditions, therefore pushing the limits of technology. The study examined the potential and constraints of existing biomarker detection capabilities by analysing UV-induced damage, spectral characteristics, pigments, and cell components over a period of 16 months. In addition to organic biomarkers, Misra et al. [5] suggested the possibility of detecting biomolecules and polyaromatic hydrocarbons from a distance, as these substances might serve as indicators of biological activities. The proposed conceptual biofinder has the potential to facilitate efficient and comprehensive scanning of large geographical areas from a satellite's orbit, with the aim of identifying possible regions of interest that need further exploration. However, the performance of the subject in issue has not yet been evaluated in the field.

Parro et al. [6] successfully identified biomarkers in analogous settings by the use of Life Marker biochips on a lower scale. The study showcased the capability of biochip technology to quantitatively assess biosignatures in their natural environment by using several biomolecular markers. Nevertheless, the performance exhibited constraints when faced with intricate field circumstances. In addition to studying organic matter, Meslin et al. [7] conducted an analysis on the geochemistry and mineralogy of soils found on Mars. Through a comparative analysis of compositions between Earth mugearites and the findings from a 100-sol observational research, valuable insights were gained on the historical water activity and magma sources on Mars. Similarly, Kounaves et al. [8] conducted a study with the objective of identifying and quantifying ionic species and heavy metals present in the regolith. The suggested methodology has the potential to enhance organic biomarker investigations by offering further evidence on historical environmental conditions. Parnell et al. [9] conducted a comprehensive assessment to identify and prioritise essential organic molecular targets for the purpose of life detection. The study suggests that in order to increase the likelihood of obtaining conclusive evidence, it is advisable to concentrate on the examination of biological and non-biological organics, meteoritic organics, fossil biomarkers, and extant biomarkers. In their study,Williford et al. [10] delineated the many scientific objectives of the Mars 2020 mission. The Mars 2020 mission demonstrates a complete methodology for detecting signs of life on Mars, which includes the examination of historical habitability, temperature, geology, as well as the direct investigation of organic molecules and microbes. In conclusion, ongoing research endeavours persistently explore the limits of biosignature detection and analysis methodologies, concurrently revealing molecular targets with significant potential to direct forthcoming endeavours in the field of life detection. Utilising a range of tactics can enhance the likelihood of achieving success in this arduous undertaking.

## III. DESIGN AND FABRICATION

Figure 1 describes a well-engineered soil collection system for a Mars rover. The system is designed to collect and analyze multiple soil samples efficiently and accurately in the harsh Martian environment. The system features a rotating plate to hold the sample, three rack and pinions to position the pH sensor, three electric actuators with pumps to collect the soil, a NEMA-17 stepper motor to drive the rotating disc, a 100mm linear actuator to control the suction pump, aluminium plates and bars to ensure stability and reduce weight, and multiple funnels and funnel covers to prevent chemical contamination. The rotating plate allows for the sequential collection and analysis of multiple samples, while the stepper motor ensures smooth and precise rotation. The linear actuator provides precise positioning of the pH sensor, while the electric actuator with pump provides powerful suction and allows operation in a variety of conditions. The 100mm linear actuator prevents overheating and extends system life, while the aluminium plates and bars provide stability and reduce weight. Multiple funnels and funnel covers prevent chemical contamination of soil samples. Overall, the soil collection system is a very well-designed and innovative solution for use on a Mars rover. It is lightweight, compact, and robust, and it features several innovative features to ensure efficient and accurate soil collection and analysis in the harsh Martian environment. In Figure 1, the full rover with sample collection and biomolecule detection sub-system is illustrated.

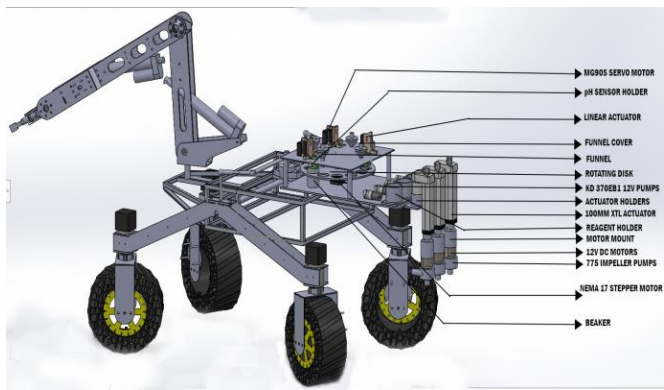

Fig. 1: Full rover with sample collection subsystem

## IV. WORKING PROCEDURE

The existence of organic molecules, the presence of liquid water, and the availability of free energy are essential prerequisites for the existence of life. Amino acids, nucleobases, and lipids are widely recognised as fundamental biomarkers found across diverse living forms, playing a crucial role in creature identification. The identification and analysis of biomarkers hold significant importance in the scientific exploration for potential life on Mars, as these organic compounds have the remarkable ability to persist across vast timescales, perhaps spanning billions of years, under suitable protective conditions. The comprehension of Earth's equivalent habitats enables the pursuit of potential life on Mars, notwithstanding the adverse climate and surface conditions it presents. Our methodology comprises three primary steps.

### A. Soil Collection

The soil collection system described uses a clearly defined process using a combination of key components. The pump is at the heart of the operation, creating a negative pressure environment inside the casing, allowing the suction of soil from the casing end. Then, the soil is transported into the casing. The impeller, driven by a DC motor and powered by a battery, rotates to create the necessary vacuum and centrifugal force, effectively sucking the soil through the tangential end of the casing body. For accurate soil collection, three electric actuators are used strategically to precisely position the pump on the soil sample. After being exploited and processed, soil will be transported through plastic pipes attached to the tangential end of the casing body to the designated location. This well-designed system provides a very efficient and effective method of soil collection, making it particularly suitable for many applications including soil analysis and research.

### B. Soil Drop

The collected soil is guided into beakers by means of a pump that expels the soil through three saline pipes. These pipes are connected to three distinct funnels positioned above the mechanism. The pump exerts sufficient pressure to propel the soil down the pipes and into the beakers, which are affixed to a spinning plate. The rotation and positioning of the plate are facilitated by a stepper motor, enabling the manipulation of the beakers to various locations as required. After the desired amount of soil has been added to a beaker, it is subsequently displaced from its original position, and a replacement beaker is introduced.

### C. Reagent & Water Supply

The reagents and water were transferred from the reagent bottles to the designated beakers containing the soil using saline pipes, facilitated by KDP-370EB 12-volt pumps. The test tubes were supplied with water in accordance with the appropriate ratio for the experiment.

### D. Analyzing Soil Test Result

The outcomes of the experiments were determined by analysing the colour alteration observed by an onboard camera and employing the Rapid multiple biomolecules based life detection methodology (MBLDPR) [3] for analysing soil samples. The samples were categorised as extinct, Extinct, or NPL (No Presence of Life).In the past, soil testing were conducted with heat-consuming methods. In order to mitigate the power consumption of the rover, all tests are transformed into heat-independent ones.

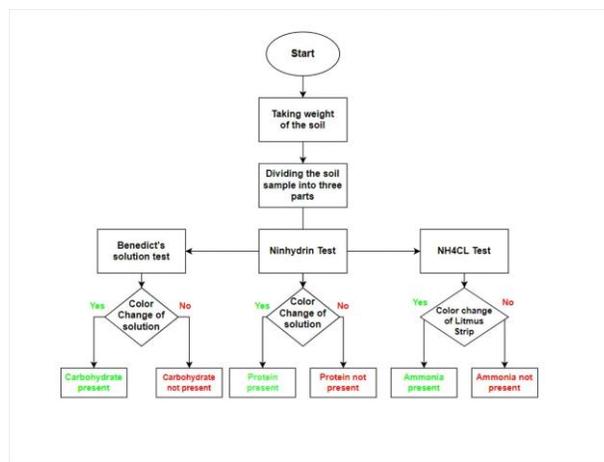

Fig. 2: Soil testing workflow diagram

## V. METHODOLOGY

The proposed methodology outlined in this study comprises a series of systematic procedures designed to facilitate extensive research and analysis in extraterrestrial environments.

We are using MBLDP-R protocol for our sub-system. The development of the Multiple Biomolecules-Based Life Detection Protocol (MBLDP-R) involved an in-depth evaluation of research carried out in four different stages. These are:

1) The compilation of a potential list of biomolecules.
2) The process of selecting biomolecules based on a thorough understanding of requirements.
3) Development of the core structure of the protocol
4) The utilisation of qualitative test scoring in selecting optimal test procedures.

Preliminary considerations for the detection protocol comprise a variety of potential biomolecules, such as carbohydrates, proteins, lipids, nucleic acids, and pigments. Nucleic acids and proteins contain genetic information. Carbohydrates are byproducts of microbial metabolism. Byproducts of bacterial nitrogen metabolism include ammonia. Pigments are derived from microbes and vegetation. Lipids are found in significant quantities within microbial and animal cells. The biomolecules in question were chosen due to their potential to serve as indicators of life's existence in the past or present via preservation, composition, or biological activity.

The biomolecules chosen for the detection process, based on a thorough review of requirements, include proteins, carbohydrates, and ammonia. The colorimetric tests commonly employed for the detection of proteins, such as the Biuret test, and carbohydrates, such as Benedict's test, are characterised by their expeditiousness, simplicity, and compatibility with rudimentary equipment that may be accommodated on the rover platform. Ammonia can be detected using readily available methods such as litmus paper and portable sensors.

The exclusion of nucleic acids is attributed to the intricate, perilous, and time-intensive extraction procedures that are deemed inappropriate for implementation on the rover. The process of pigment detection encompasses a series of sequential steps, such as chromatography, which is not feasible to be conducted by the rover. The measurement of lipids necessitates the utilisation of solvent extraction techniques as well as gravimetric and chromatographic tests, which pose challenges in terms of remote implementation.

The selected biomolecules and testing methodologies satisfy essential characteristics such as rapidity, ease of use, portability of equipment, and adherence to the resource limitations and competition prerequisites of the rover. The requirement analysis conducted in this study aims to assess the feasibility of the detection process under the specified operating circumstances and restrictions.

For detection, a three-level decision tree [4] was utilised. Protein presence indicates the existence of life, carbohydrate presence in the absence of protein indicates the extinction of life, and ammonium content alone is inconclusive; therefore, life is absent. Protein is critical for modern life, whereas persistent carbohydrates indicate the existence of life on prehistoric rocks. The classification of samples into Extant, Extinct, or No Life is facilitated by the decision tree by these critical biomolecules. This facilitates the investigation process and eradicates any potential confusion. Figure 3 illustrates overall workflow of our sub-system.

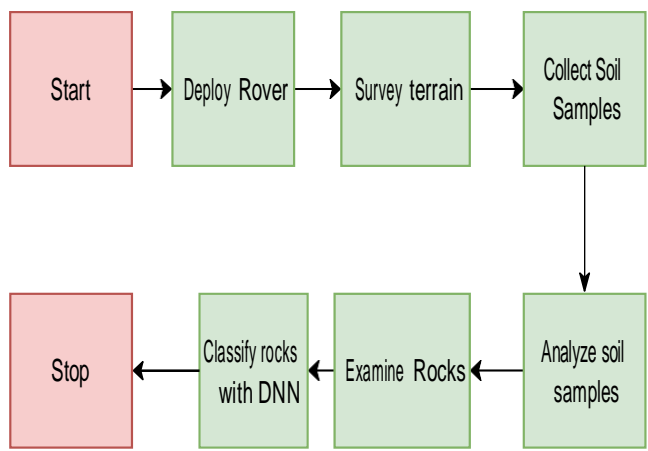

Fig. 3: Overall Workflow

The methodology consists of eighteen key steps, beginning with the deployment of a rover at the designated exploration site. Following this, the cameras and sensors installed on the vehicle are utilised to carry out surveys of the terrain and accurately identify specific regions that are of relevance for sampling. The procedure starts by initiating the soil extraction equipment, thereafter proceeding to evaluate the initial pH levels at the designated sample location. The collection of soil samples involves the utilisation of a suction pump to extract the soil, which is then carefully placed into specifically designated containers. This process is repeated multiple times to gather samples from different depths and locations, employing an iterative suction sampling technique. The operation additionally involves the rotation of sample plates, the repositioning of the rover for supplementary soil collection, and the subsequent transportation of soil samples to an onboard analysis station. Chemical reagents are used to trigger reactions in the samples, and the resulting colour changes are captured and recorded using camera imaging techniques. The acquired data is subsequently employed to categorise soil samples according to test outcomes, differentiating between present life, past life, or the lack thereof. The process includes the examination of rocks, wherein sensors and microscopes are strategically placed over specific rocks to facilitate imaging. Data is collected from multiple pieces of equipment to identify the presence of microorganisms.

A Deep Neural Network (DNN) model is utilised to categorise rock samples, distinguishing between fossilised specimens and those that are not. The aforementioned comprehensive methodology is consistently employed in the examination and categorization of numerous geological formations for the goals of surveying and classification. In the end, the primary outcomes and representative data are communicated to the ground control station for subsequent analysis, enhancing the comprehensive and rigorous scientific investigation procedure.

Our DNN implementation centred around the detection of igneous/metamorphic rock and shale, which are geological elements present on the Martian surface and potentially harbour indications of life. To accomplish this, we utilised the VGG16 model, which is a convolutional neural network renowned for its efficacy in the domain of image classification. We were able to contribute to the comprehension of Martian geology and the search for potential indicators of extraterrestrial life by employing this methodology.

**Algorithm 1** Bio-Sensor Data Collection

1: Initialize Serial Monitor at 9600 baud
2: Set sensor pins as INPUT/OUTPUT
3: Configure color sensor frequency scaling
4: Configure GASpin, ALCpin, FDpin, NH3pin as INPUT
5: Wait 2 seconds for sensor setup
6: **while** True **do**
7:    Read Red, Green, Blue values from color sensor
8:    Map and convert values to 0-255 range
9:    Print RGB values to Serial Monitor
10:    **Read alcohol sensor** digital output
11:    Print if alcohol detected or not
12:    Read raw analog value **from CO2 sensor**
13:    Calculate and print CO2 level (ppm)
14:    Read analog voltage from **formaldehyde sensor**
15:    Calculate and print concentration (ppm)
16:    Read analog **humidity value**
17:    Calculate & print humidity percentage
18:    Measure voltage drop across **ammonia sensor**
19:    Calculate Rs/Ro ratio
20:    Print calculated ammonia level (ppm)
21: **end while**

### A. Biosensor Mechanism and Algorithm

The bio sensor technique utilised in this study employs colorimetric assays using paper strips that are coated with certain reagents. For instance, Benedict's solution is used for detecting carbohydrates, Ninhydrin solution is used for detecting proteins, and Nessler's reagent is used for detecting ammonia. The application of soil samples onto designated strips is followed by the occurrence of a colour shift, indicating the presence of the specific biomolecule of interest. This colour change facilitates the estimation of concentration levels through reference to a colour chart. Biological sensors have several benefits, including their straightforwardness, rapidity, cost-effectiveness, mobility, and adaptability in the identification of diverse biomolecules. The biosensors utilised in this research study are referenced in table I.

Nevertheless, it is worth noting that these alternative methods may demonstrate reduced sensitivity in comparison to conventional techniques, as well as a heightened vulnerability to influence from soil constituents such as humic acids and metal ions. In brief, bio sensors offer a handy method for detecting biomolecules in soil; nonetheless, it is important to acknowledge the limitations of sensitivity and potential interference. The algorithm used for this study is shown in algorithm 1.

| Sensor Name | Purpose |
|---|---|
| MQ137 | Detects ammonia released by microorganisms |
| MQ135 | Detects CO2 released by microorganisms |
| MQ3 | Detects alcohol levels on rock surfaces |
| MQ138 | Detects formaldehyde levels on rocks |
| TCS3200 | Detects color of rocks to identify rock type |
| YL-69 | Analyzes soil moisture content |
| HR202 | Detects air humidity to analyze habitability |
| Analog pH sensor | Detects pH levels in soil samples |
| USB microscope | Examines rock samples at microscopic level |
| 1080p camera | Captures images of samples for analysis |
| mVOC sensors | Detects gases indicating microorganisms |

TABLE I: BioSensor Components

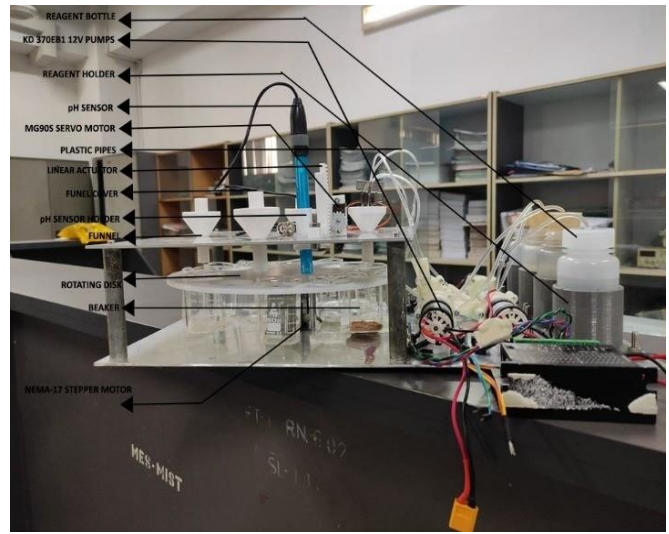

Fig. 4: Soil Analysis subsystem on rover

## VI. RESULT ANALYSIS

### A. Mechanical Improvement Analysis

Our methodology has improved the following mechanical aspects :

- Increased depth of soil collection.
- Strict sterilization procedures.
- Supplementation of bio loads.
- Use of a NEMA-17 stepper motor for sample container control.
- Use of a 3D printed pH sensor with linear actuator.
- Streamlined production and reduced costs due to the use of 3D printed components.
- Use of NEMA-17 stepper motors for rotation and 12V DC motors for uninterrupted rotation.
- An actuator operation cycle of 2 minutes on and 18 minutes off.
- A unified platform for system component consolidation.
- Multiple effective funnels and covers for chemical contamination.
- Compact and lightweight soil collector design

| Serial No. | Aspect | Metrics |
|---|---|---|
| 1 | Depth of Soil Collection | > 5 cm |
| 2 | MSterilization Procedures | Strict |
| 3 | Bio Loads Supplemented | Dextrose, Albumin, Ammonia |
| 4 | Motor for Sample Container Control | NEMA-17 Stepper Motor |
| 5 | pH Sensor with Linear Actuator | 3D Printed |
| 6 | 3D Printed Components | Streamlined Production, Reduced Costs |
| 7 | Motor Selection for Precise Movement | NEMA-17 Stepper Motors (Rotation), 12V DC Motors (Uninterrupted Rotation) |
| 8 | Actuator Operation Cycle | 2 minutes on, 18 minutes off |
| 9 | System Component Consolidation | Unified Platform |
| 10 | Funnels and Covers for Chemical Contamination | Multiple, Effective |
| 11 | Soil Collector Design | Compact and Lightweight |

TABLE II: Improved Features

*B. Bio-molecular analysis*

Sensitivity evaluation of biomolecule tests found from this study are:

1) **Benedict's reagent:** Sensitivity to carbohydrates of a high order.
2) **Ninhydrin test with smaller samples (2-5 grams):** 7-minute reaction time, inaccurate negative results.
3) **Protein detection:** 5 minutes, 10 grams of samples, 20 ml. of reagents.
4) **Ammonium ion test:** Can be detected in 3 minutes.

The results of this research underscore the significance of improving the Ninhydrin analysis for trace proteins, recognising the necessity for considerable sample size and supplementary reagents when dealing with low-concentration biomarkers, and recognising the dependability of chemical assays despite fluctuations in parameters.

## VII. CONCLUSION AND FUTURE WORK

Our research introduces a cost-effective mobile robot with a specialized system for analyzing biomolecules to search for extraterrestrial life. The upgraded Phoenix rover possesses precise imaging and sampling capabilities that augment the analysis of soil and rock samples. Field trials validate its capability to traverse diverse terrains and gather samples for the purpose of biomarker analysis. Investing in long-lasting carbon-based organic compounds, which are renowned for their resilience in geological environments, exhibits potential. More rigorous testing under conditions resembling those of Mars is advised. Subsequent investigations ought to expand the spectrum of biomolecules that can be detected, enhance the capabilities of machine learning, and amass a varied corpus of training data. This prospective interdisciplinary approach, which integrates robotics and analytical chemistry, has the capacity to facilitate future astrobiology missions and advance the search for extraterrestrial life.